\documentclass[sigconf]{acmart}

\usepackage{natbib}
\usepackage{amsmath,amsfonts}
\usepackage{graphicx}
\usepackage{textcomp}
\usepackage{xcolor}
\def\BibTeX{{\rm B\kern-.05em{\sc i\kern-.025em b}\kern-.08em
    T\kern-.1667em\lower.7ex\hbox{E}\kern-.125emX}}

\usepackage{amsmath}
\usepackage{algorithm}
\usepackage{enumitem}
\usepackage{picinpar}

\usepackage{algorithmicx}
\usepackage[noend]{algpseudocode}
\usepackage{subfigure}
\usepackage{multirow}
\usepackage{color}
\usepackage{enumitem}
\usepackage{hhline}
\usepackage[normalem]{ulem}
\usepackage{booktabs}
\usepackage{wrapfig}
\usepackage{cancel}
\usepackage{hyperref}

\newcommand{\bL}{\ensuremath{\mathcal{L}}}

\renewcommand{\vec}[1]{\ensuremath{\mathbf{#1}}}

\newcommand{\stitle}[1]{\vspace{1mm} \noindent {\bf #1}}

\newcommand{\eg}{{\it e.g.}}

\newcommand{\ie}{{\it i.e.}}
\newcommand{\etc}{{\it etc.}}

\newcommand{\method}[1]{{#1}}
\newcommand{\model}{\method{LHGNN}{}}

\newcommand{\eat}[1]{}

\newcommand{\stkout}[1]{\ifmmode\text{\sout{\ensuremath{#1}}}\else\sout{#1}\fi}

\AtBeginDocument{%
  \providecommand\BibTeX{{%
    \normalfont B\kern-0.5em{\scshape i\kern-0.25em b}\kern-0.8em\TeX}}}

\copyrightyear{2023}
\acmYear{2023}
\setcopyright{acmlicensed}\acmConference[WWW '23]{Proceedings of the ACM Web Conference 2023}{May 1--5, 2023}{Austin, TX, USA} \acmBooktitle{Proceedings of the ACM Web Conference 2023 (WWW '23), May 1--5, 2023, Austin, TX, USA}
\acmPrice{15.00}
\acmDOI{10.1145/3543507.3583284} \acmISBN{978-1-4503-9416-1/23/04}

\title{Link Prediction on Latent Heterogeneous Graphs}

\begin{document}

\author{Trung-Kien Nguyen$^*$}
\affiliation{
  \institution{Singapore Management University}
  \country{Singapore}
  }
\email{tknguyen@smu.edu.sg}

\author{Zemin Liu$^{*\dagger}$}
\affiliation{
 \institution{National University of Singapore}
 \country{Singapore}
 }
\email{zeminliu@nus.edu.sg}

\author{Yuan Fang}
\affiliation{
 \institution{Singapore Management University}
 \country{Singapore}
 }
\email{yfang@smu.edu.sg}

\thanks{
    $^*$Co-first authors with equal contribution.
    \\$^{\dagger}$Corresponding author. Work partly done at Singapore Management University.
}

\renewcommand{\shortauthors}{Trung-Kien Nguyen, Zemin Liu, and Yuan Fang}

\begin{abstract}
 On graph data, the multitude of node or edge types gives rise to heterogeneous information networks (HINs). To preserve the heterogeneous semantics on HINs, the rich node/edge types become a cornerstone of HIN representation learning. However, in real-world scenarios, type information is often noisy, missing or inaccessible. Assuming no type information is given, we define a so-called \emph{latent heterogeneous graph} (LHG), which carries latent heterogeneous semantics as the node/edge types cannot be observed. 
In this paper, we study the challenging and unexplored problem of link prediction on an LHG. As existing approaches
depend heavily on type-based information, they are suboptimal or even inapplicable on LHGs. To address the absence of type information, we propose a model named \model, based on the novel idea of semantic embedding at node  and path levels, to capture latent semantics on and between nodes.
We further design a personalization function to modulate the heterogeneous contexts conditioned on their latent semantics w.r.t.~the target node, to enable finer-grained aggregation. 
Finally, we conduct extensive experiments on four benchmark datasets, and demonstrate the superior performance of \model.
\end{abstract}



\maketitle

\section{Introduction} \label{sec.intro}

\begin{figure}[t]
\centering
\includegraphics[width=0.99\linewidth]{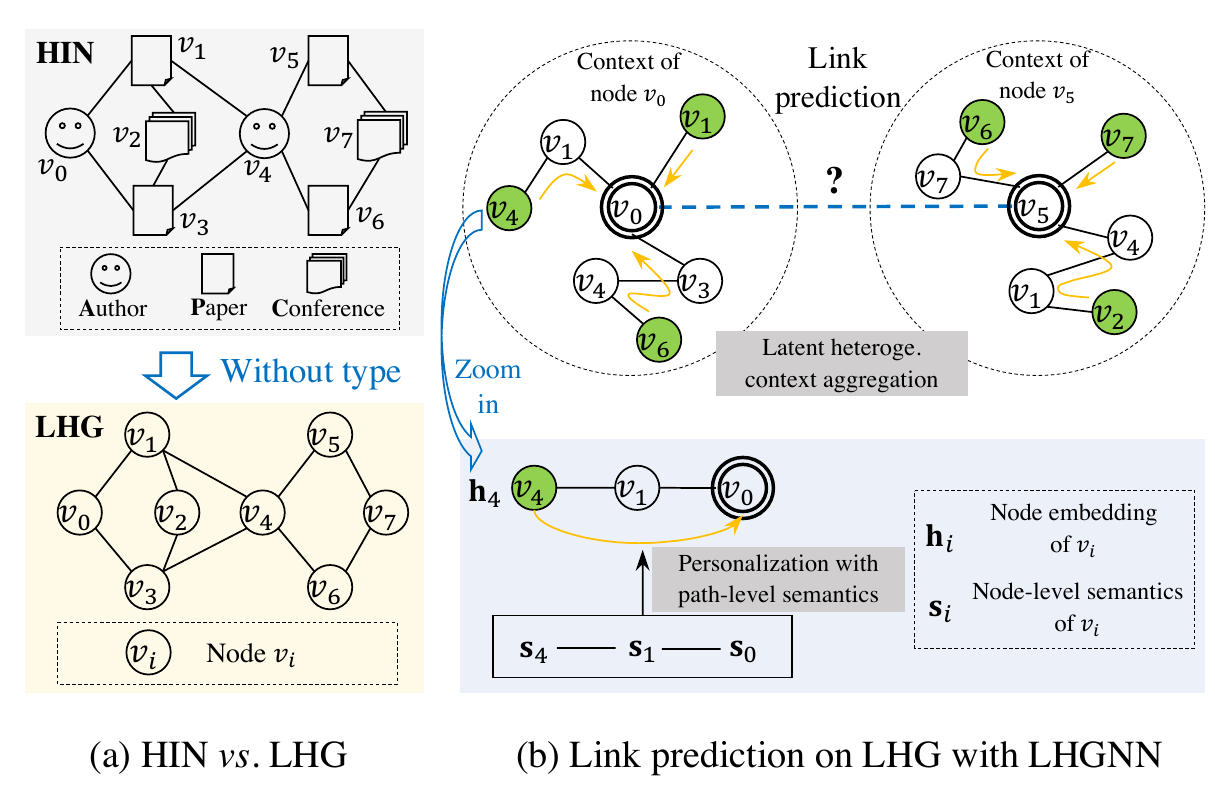}
\vspace{-3mm}
\caption{Illustration of our problem and approach. (a) Comparison of HIN and LHG. (b) Key insights of our approach.}
\label{fig.intro-motivation}
\end{figure}

Objects often interact with each other to form graphs, such as the Web and social networks. The prevalence of graph data has catalyzed graph analysis in various disciplines. In particular,
link prediction \cite{zhang2018link} is a fundamental graph analysis task, enabling widespread applications such as friend suggestion in social networks, recommendation in e-commerce graphs, and citation prediction in academic networks. 
In these real-world applications, the graphs are typically \emph{heterogeneous} as opposed to \emph{homogeneous}, also known as Heterogeneous Information Networks (HINs) \cite{yang2020heterogeneous}, in which multiple types of nodes and edges exist. 
For instance, the academic HIN shown in the top half of Fig.~\ref{fig.intro-motivation}(a) interconnects nodes of three types, namely, Author (A), Paper (P) and Conference (C), through  different types of edges such as ``writes/written by'' between author and paper nodes and ``publishes/published in'' between conference and paper nodes, \etc\
The multitude of node and edge types in HINs implies rich and diverse semantics on and between nodes, which opens up great opportunities for link prediction.  

A crucial step for link prediction is to derive features from an input graph.
Recent literature focuses on graph representation learning \cite{cai2018comprehensive,wu2020comprehensive}, which aims to map the nodes on the graph into a low-dimensional space that preserves the graph structures.
Various approaches exist, ranging from the earlier shallow embedding models \cite{perozzi2014deepwalk,tang2015line,grover2016node2vec}
to more recent message-passing graph neural networks (GNNs) \cite{gcn,gat,sage,xu2018powerful}.
Representation learning on HINs generally follows the same paradigm, but aims to preserve the heterogeneous semantics in addition to the graph structures in the low-dimensional space. To express heterogeneous semantics, existing work resorts to type-based information, including simple node/edge types (\eg, an author node carries different semantics from a paper node), and type structures like metapath \cite{sun2011pathsim} (\eg, the metapath A-P-A implies two authors are collaborators, whereas A-P-C-P-A implies two authors in the same field; see  Sect.~\ref{sec:prelim} for the metapath definition).
Among the state-of-the-art heterogeneous GNNs, while hinging on the common operation of message passing, some exploit node/edge types \cite{hgt,hgn,hetgnn,hong2020attention} and others employ  type structures \cite{han,fu2020magnn,sankar2019meta}.

\stitle{Our problem.}
The multitude of node or edge types gives rise to rich heterogeneous semantics on HINs, and forms the key thesis of HIN representation learning \cite{yang2020heterogeneous}.  
However, in many real-world scenarios, type information is often noisy, missing or inaccessible.
One reason is type information does not exist explicitly and has to be deduced. For instance, when extracting entities and their relations from texts to construct a knowledge graph, NLP techniques are widely used to classify the extractions into different types, which can be noisy.
Another reason is privacy and security, such that the nodes in a network may partially or fully hide their identities and types.
Lastly, even on an apparently homogeneous graph, such as a social network which only consists of users and their mutual friendships, could have fine-grained latent types, such as different types of users (\eg, students and professionals) and different types of friendships (\eg, friends, family and colleagues), but we cannot observe such latent types.

Formally, we call those HINs without explicit type information as \emph{Latent Heterogeneous Graphs} (LHGs), as shown in the bottom half of  Fig.~\ref{fig.intro-motivation}(a). 
The key difference on an LHG is that, while different types still exist, the type information is completely inaccessible and cannot be observed by data consumers.
It implies that LHGs still carry rich heterogeneous semantics that are crucial to effective representation learning, but the heterogeneity becomes latent and presents a much more challenging scenario given that types or type structures can no longer be used. 
In this paper, we investigate this unexplored problem of \emph{link prediction on LHGs}, which calls for modeling the latent semantics on LHGs as links are formed out of relational semantics between nodes.

\stitle{Challenges and insights.}
We propose a novel model for link prediction on LHGs, named \underline{L}atent \underline{H}eterogeneous \underline{G}raph \underline{N}eural \underline{N}etwork (\model).
Our general idea is to develop a latent heterogeneous message-passing mechanism on an LHG, in order to exploit the latent semantics between nodes for link prediction.
More specifically, we must address two major challenges.

First, 
\emph{how to capture the latent semantics on and between nodes without any type information?}
In the absence of explicit type information, we resort to \emph{semantic embedding} at both node and path levels to depict the latent semantics. At the node level, we complement the traditional  representation of each node (\eg, $\vec{h}_4$ in Fig.~\ref{fig.intro-motivation}(b), which we call the \emph{primary embedding}) with an additional semantic embedding (\eg, $\vec{s}_4$). While the primary embedding can be regarded as a blend of content and structure information, the semantic embedding aims to distill the more subtle semantic information (\ie, latent node type and relation types a node tends to associate with), which could have been directly expressed if explicit types were available. Subsequently, at the path level, we can learn a semantic path embedding based on the node-level semantic embeddings in a path on an LHG, \eg, $\vec{s}_4$, $\vec{s}_1$ and $\vec{s}_0$ in the path $v_4$--$v_1$--$v_0$ in Fig.~\ref{fig.intro-motivation}(b). The path-level semantic embeddings aim to capture the latent high-order relational semantics between nodes, such as the collaborator relation between authors $v_4$ and $v_0$ in Fig.~\ref{fig.intro-motivation}(a), to mimic the role played by metapaths such as A-P-A.

Second, as context nodes of a target node often carry latent heterogeneous semantics, \emph{how to differentiate them for finer-grained context aggregation?}
We propose a learnable \emph{personalization} function to modulate the original message from each context node. The personalization hinges on the semantic path embedding between each context node and the target node as the key differentiator of heterogeneous semantics carried by different context nodes.  
For illustration, refer to the top half of Fig.~\ref{fig.intro-motivation}(b), where the green nodes (\eg, $v_1$, $v_4$ and $v_6$) are the context nodes of the doubly-circled target node (\eg, $v_0$). These context nodes carry latent heterogeneous semantics (\eg, $v_1$ is a paper written by $v_0$, $v_4$ is a collaborator of $v_0$, and $v_6$ is a related paper $v_0$ might be interested in), and thus can be personalized by the semantic path embedding between each context and the target node before aggregating them. 


\stitle{Contributions.}
In summary, our contributions are three-fold.
(1) We investigate a novel problem of link prediction on latent heterogeneous graphs, which differs from traditional HINs due to the absence of type information.
(2) We propose a novel model \model\ based on the key idea of semantic embedding to bridge the gap for representation learning on LHGs. \model\ is capable of inferring both node- and path-level semantics, in order to personalize the latent heterogeneous contexts for finer-grained message passing within a GNN architecture.
(3) Extensive experiments on four real-world datasets demonstrate the superior performance of \model\ in comparison to the state-of-the-art baselines.

\section{Related Work}


\stitle{Graph neural networks.} 
GNNs have recently become the mainstream for graph representation learning.
Modern GNNs typically follow a message-passing scheme, which derives low-dimensional embedding of a target node by aggregating messages from context nodes. Different schemes of context aggregation have been proposed, ranging from simple  mean pooling \cite{gcn,sage} to neural attention \cite{gat} and other neural networks \cite{xu2018powerful}. 

For representation learning on HINs, a plethora of heterogeneous GNNs have been proposed. They depend on type-based information to capture the heterogeneity, just as earlier HIN embedding approaches \cite{metapath2vec,fu2017hin2vec,hu2019adversarial}.
On one hand, many of them leverage simple node/edge types.   
HetGNN \cite{hetgnn} groups random walks by node types, and then applies bi-LSTM to aggregate node- and type-level messages. HGT \cite{hgt} employs node- and edge-type dependent parameters to compute the heterogeneous attention over each edge.
Simple-HGN \cite{hgn} extends the edge attention with a learnable edge type embedding, whereas HetSANN \cite{hong2020attention} employs a type-aware attention layer.
On the other hand, high-order type structures such as meta-path \cite{sun2011pathsim} or meta-graph \cite{fang2016semantic} have also been used. 
HAN \cite{han} uses meta-paths to build homogeneous neighbor graphs to facilitate node- and semantic-level attentions in message aggregation, whereas MAGNN \cite{fu2020magnn} proposes several meta-path instance encoders to account for intermediate nodes in a meta-path instance. Meta-GNN \cite{sankar2019meta} differentiates context nodes based on meta-graphs. 
Another work Space4HGNN \cite{space4hgnn} proposes a unified design space to build heterogeneous GNNs in a modularized manner, which can potentially utilize various type-based information.
Despite their effectiveness on HINs, they cannot be applied to LHGs due to the need of explicit type information. 


\stitle{Knowledge graph embedding.}
A knowledge graph consists of a large number of relations between head and tail entities. 
Translation models \cite{transe,wang2014knowledge,transr} are popular approaches that treat each relation as a translation in some vector space. For example, TransE \cite{transe} models a relation as a translation between head and tail entities in the same embedding space, while TransR \cite{transr} further maps entities into multiple relation spaces to enhance semantic expressiveness. Separately, RotatE \cite{rotate} models each relation as a rotation from head to tail entities in complex space, and is able to capture different relation patterns such as symmetry and inversion. These models require the edge type (\ie, relation) as input, which is not available on an LHG. Compared to heterogeneous GNNs, they do not utilize entity features or model multi-hop interactions between entities, which can lead to inferior performance on HINs.

\stitle{Predicting missing types.}
A few studies \cite{neelakantan2015missing,hovda2019missing} aim to predict missing entity types on knowledge graphs. However, a recent work \cite{zhang2021relation} shows that these approaches tend to propagate type prediction errors on the graph, which harms the performance of other tasks like link prediction. Therefore, RPGNN \cite{zhang2021relation} proposes a relation encoder, which is adaptive for each node pair to handle the missing types. However, all of them still require partial type information from a subset of nodes and edges for supervised training, which makes them infeasible in the LHG setting.

\section{Preliminaries}\label{sec:prelim}

We first review or define the core concepts in this work.


\stitle{Heterogeneous Information Networks (HIN).} An HIN \cite{sun2012mining} is defined as a graph $G=(V,E,T,R,\psi, \phi)$, where $V$ denotes the set of nodes and
$T$ denotes the set of node types, $E$ denotes the set of edges and $R$ denotes the set of edge types. Moreover, $\psi:V\to T$ and $\phi:E\to R$ are functions that map each node and edge to their types in $T$ and $R$, respectively. $G$ is an HIN if $|T|+|R|>2$.

\stitle{Latent Heterogeneous Graph (LHG).} 
An LHG is an HIN $G=(V,E,T,R, \psi, \phi)$ such that the types $T,R$ and mapping functions $\psi,\phi$ are \emph{not accessible}. That is, we only observe a \emph{homogeneous} graph $G'=(V,E)$ without knowing $T,R,\psi,\phi$.

\stitle{Metapath and latent metapath.} On an HIN, a metapath $P$ is a sequence of node and edge types \cite{sun2011pathsim}: $P=T_{1} \xrightarrow[]{R_{1}} T_{2} \xrightarrow[]{R_{2}}...\xrightarrow[]{R_{l}}T_{l+1}$, such that $T_{i}\in T$ and $R_{i} \in R$. As an edge type $R_i \in R$ represents a relation, a metapath represents a composition of relations $R_1 \circ R_2 \circ \ldots \circ R_l$.
Hence, metapaths can capture complex, high-order semantics between nodes.
A path $p=(v_1,v_2,\ldots,v_{l+1})$ on the HIN is an instance of metapath $P$ if
$\psi(v_i)=T_i$ and $\phi(\langle v_i,v_{i+1}\rangle)=R_i$. 
As shown in the top half of Fig.~\ref{fig.intro-motivation}(a), an example metapath is Author-Paper-Author (A-P-A), implying the collaborator relation between authors. Instances of this metapath include $v_0$-$v_1$-$v_4$ and $v_0$-$v_3$-$v_4$, signifying that $v_0$ and $v_4$ are collaborators.

On an LHG the metapaths become latent too, as the types $T_i$ and $R_i$ are not observable. Generally, any path $p=(v_1,v_2,\ldots,v_{l+1})$ on an LHG is an instance of some latent metapath, which carries latent semantics representing an unknown composition of relations between the starting node $v_1$ and end node $v_{l+1}$.

\stitle{Link prediction on LHG.} Given a query node, we rank other nodes by their probability of forming a link with the query. The difference lies in the input graph, where we are given an LHG.




\begin{figure*}[t]
\centering
\includegraphics[width=0.99\linewidth]{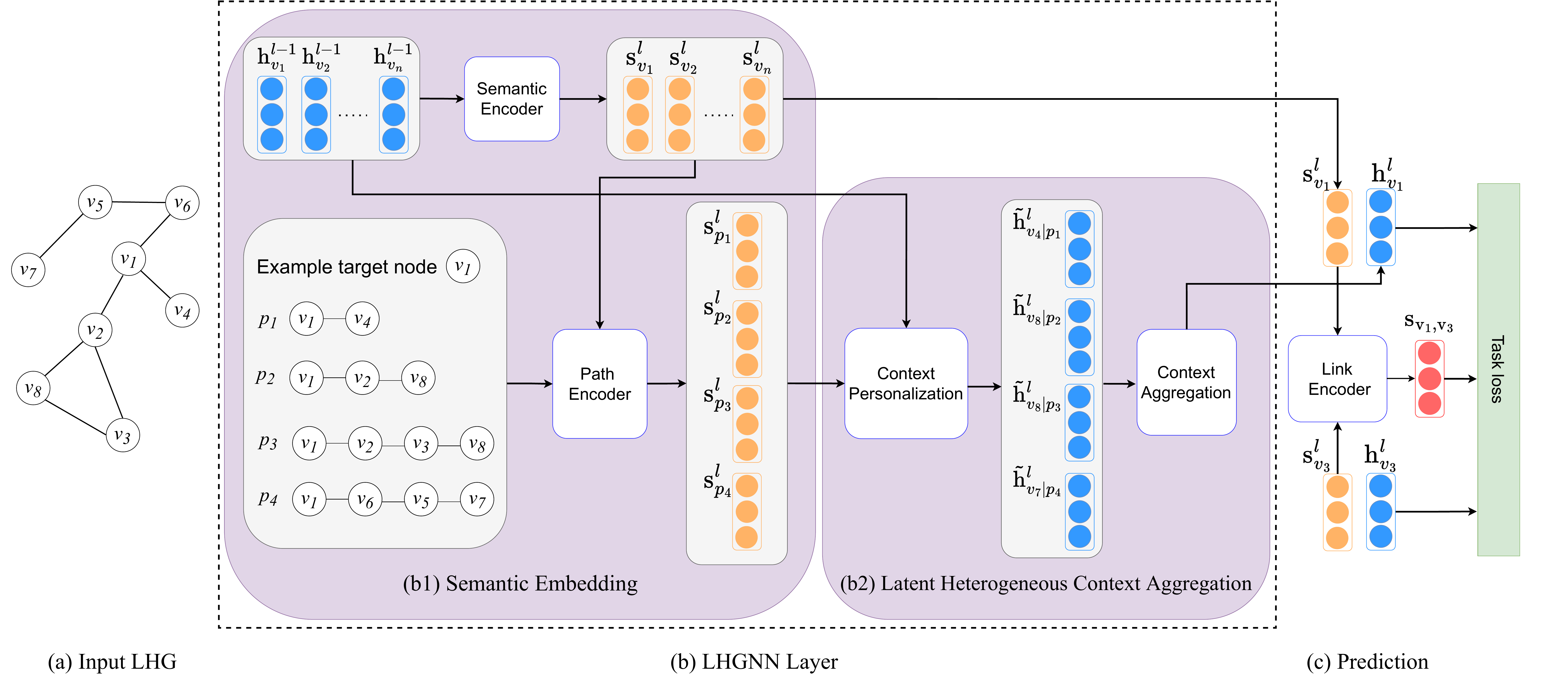}
\vspace{-2mm}
\caption{Overall framework of \model.}
\label{fig.model}
\end{figure*}

\section{Proposed Method: \model}

In this section, we introduce the proposed method \model\ for link prediction on latent heterogeneous graphs. 

\subsection{Overall Framework}

We start with the overall framework of \model, as presented in Fig.~\ref{fig.model}.
An LHG is given as input as illustrated in Fig.~\ref{fig.model}(a), which is fed into an \model~layer in Fig.~\ref{fig.model}(b). Multiple layers can be stacked, and the last layer would output the node representations, to be further fed into a link encoder for the task of link prediction as shown in Fig.~\ref{fig.model}(c). 
More specifically, the \model~layer is our core component, which consists of two sub-modules: a \emph{semantic embedding} sub-module to learn node-level and path-level latent semantics, and a \emph{latent heterogeneous context aggregation} sub-module to aggregate messages for the target node. We describe each sub-module and the link prediction task in the following.

\subsection{Semantic Embedding}

Semantic embedding aims to model both node- and path-level latent semantics, as illustrated in Fig.~\ref{fig.model}(b1).

\stitle{Node-level semantic embedding.} 
For each node $v$, alongside its primary embedding $\mathbf{h}_v$, we propose an additional semantic embedding $\mathbf{s}_v$. Similar to node embeddings on homogeneous graphs, the primary embeddings intend to capture the overall content and structure information of nodes. However, on an LHG, the content of a node contains not only concrete topics and preferences, but also subtle semantic information inherent to nodes of each latent type (\eg, node type, and potential single or multi-hop relations a node tends to be part of). Hence, we propose a semantic encoder to locate and distill the subtle semantic information from the primary embeddings to generate semantic embeddings, which will be later employed for link prediction. Note that this is different from disentangled representation learning \cite{bengio2013representation}, which can be regarded as disentangling a mixture of latent topics.

Specifically, in the $l$-th layer, given primary embeddings from the previous layer\footnote{When $l=0$, the primary embedding $\vec{h}_{v_i}^{0}$ is set to the input node features of $v_i$.} ($\vec{h}_{v_1}^{l-1}, \vec{h}_{v_2}^{l-1},\ldots$), a semantic encoder generates the corresponding semantic embeddings ($\vec{s}_{v_1}^{l},\vec{s}_{v_2}^{l},\ldots$). For each node $v$, 
the semantic encoder $f_s$ extracts the semantic embedding $\vec{s}^{l}_v$ from its primary embedding $\vec{h}^{l-1}_v$: $\vec{s}^{l}_v=f_s(\vec{h}^{l-1}_v;\theta^{l}_s)\in\mathbb{R}^{d_s^{l}}$. While the function $f_s(\cdot;\theta^{l}_s)$ can take many forms, we simply materialize it as a fully connected layer: 
\begin{align}
\vec{s}^{l}_v=\textsc{LeakyReLU}(\vec{W}^{l}_s\vec{h}^{l-1}_v+\vec{b}^{l}_s),    \label{eq:semantic-embedding-node}
\end{align}
where $\vec{W}^{l}_s\in\mathbb{R}^{d_s^{l}\times d_h^{l-1}}$ and $\vec{b}^{l}_s\in\mathbb{R}^{d_s^{l}}$ are the parameters of the encoder, \ie, $\theta^l_s=\{\vec{W}^{l}_s, \vec{b}^{l}_s\}$. Since the semantic embedding only distill the subtle semantic information from the primary embedding, it needs much fewer dimensions, \ie, $d_s^{l} \ll d_h^{l}$. 

 \stitle{Path-level semantic embedding.}
 A target node is often connected with many context nodes through paths. On an LHG, these paths may carry different latent semantics by virtue of the heterogeneous multi-hop relations between nodes.
In an HIN, to capture the heterogeneous semantics from different context nodes, metapaths have been a popular tool. For example, in the top half of Fig.~\ref{fig.intro-motivation}(a), the HIN consists of three types of nodes. There exist different metapaths that capture different semantics between authors: A-P-A for authors who have directly collaborated, or A-P-C-P-A for authors who are likely in the same field, \etc\ However, on an LHG, we do not have access to the  node types, and thus are unable to define or use any metapath. 
Thus, we employ a path encoder to fuse the node-level semantic embeddings associated with a path into a path-level embedding. The path-level semantic embeddings attempt to mimic the role of metapaths on an HIN, to capture the latent heterogeneous semantics between nodes.

 Concretely, we first perform random walks to sample a set of paths. Starting from each target node, we sample $N$ random walks of length $L_{\text{max}}$, \eg, $\{p_1,\ldots,p_4\}$ for $v_1$ as shown in Fig.~\ref{fig.model}(b).
 Then, for each sampled walk, we truncate it into a shorter path with length $L\le L_{\text{max}}$. These paths of varying lengths can capture latent semantics of different ranges, and serve as the instances of different latent metagraphs. 
 
 Next, for each path $p_i$, a path encoder encodes it to generate a semantic path embedding $\vec{s}_{p_i}^l$. 
Let $P_v$ denote the set of sampled paths starting from a target node $v$. If there exists a path $p_i \in P_v$ such that $p_i$ ends at node $u$, we call $u$ a \emph{context node} of $v$ or simply \emph{context} of $v$.
For instance, given the set of paths $P_{v_1}=\{p_1,\ldots,p_4\}$ for the target node $v_1$ in Fig.~\ref{fig.model}(b1), the contexts are $v_4,v_8,v_7$, which carry different semantics for $v_1$.
%
%
%
In the $l$-th layer, we employ a path encoder $f_p$ to embed each path $p_i$ into a semantic path embedding  $\vec{s}^{l}_{p_i}\in\mathbb{R}^{d_s^{l}}$ based on the node-level semantic embeddings of each node $v_j$ in the path. 
\begin{align}
    \vec{s}^{l}_{p_i} = f_p(\{\vec{s}_{v_j}^{l} \mid v_j \text{ in the path }  p_i\}), \label{eq:semantic-embedding-path}
\end{align}
where $f_p$ can take many forms, ranging from simple pooling to recurrent neural networks \cite{nallapati2017summarunner} and transformers \cite{vaswani2017attention}. As the design of path encoder is not the focus of this paper, we simply implement it as a mean pooling, which computes the mean of the node-level semantic embeddings as the path-level embedding.

\textsc{\textbf{Remark.}}
It is possible or even likely to have multiple paths between the target and context nodes. Intrinsically, these paths are instances of one or more  latent metapaths, which bear a two-fold advantage. 
First, as each latent metapath depicts a particular semantic relationship between two nodes, having plural latent metapaths could capture different semantics between the two nodes. This is more general than many heterogeneous GNNs \cite{han,fu2020magnn} on a HIN, which rely on a few handcrafted metapaths.     
Second, it is much more efficient to sample and process paths than subgraphs \cite{jiang2021pre}. Although a metagraph can express more complex semantics than a metapath \cite{DBLP:journals/tkde/ZhangFLWZ22}, the combination of multiple metapaths is a good approximation especially given the efficiency trade-off. 

\subsection{Latent Heterogeneous Context Aggregation} \label{sec.agg}

To derive the primary embedding of the target node, the next step is to perform latent heterogeneous context aggregation for the target node. The aggregation generally follows a message-passing GNN architecture, where messages from context nodes are passed to the target node. 
For example, consider the target node $v_1$ shown in Fig.~\ref{fig.model}(b2). The contexts of $v_1$ include $\{v_4,v_8,v_7\}$ based on the paths $p_1,\ldots,p_4$, and their messages (i.e., embeddings) would be aggregated to generate the primary embedding of $v_1$.

However, on an LHG, the contexts of a target node carry latent heterogeneous semantics. Thus, these heterogeneous contexts shall be differentiated before aggregating them, in order to preserve the latent semantics in fine granules.
Note that on an HIN, the given node/edge types or type structures can be employed as explicit differentiators for the contexts. In contrast, the lack of type information on  an LHG prevents the adoption of the explicit differentors, and we resort to path semantic embeddings as the basis to differentiate the messages from heterogeneous contexts.
That is, in each layer, we \emph{personalize} the message from each context node conditioned on the semantic path embeddings between the target and context node. The personalized context messages are finally aggregated to generate the primary embedding of the target node, \ie, the output of the current layer.

\stitle{Context personalization.}
Consider a target node $v$, a context node $u$, and their connecting path $p$.
In particular, $u$ can be differentiated from other contexts of $v$ given their connecting path $p$, which is associated with a unique semantic path embedding.
Note that there may be more than one paths connecting $v$ and $u$. In that case, we treat $u$ as multiple contexts, one instance for each path, as each path carries different latent semantics and the corresponding context instance needs to be differentiated.

Specifically, in the $l$-th layer, we personalize the message from $u$ to $v$ with a learnable transformation function $\tau$, which modulates $u$'s original message $\vec{h}^{l-1}_u$ (i.e., its primary embedding from the previous layer) into a personalized message $\tilde{\vec{h}}^{l}_{u|p}$ conditioned on the path $p$ between $u$ and $v$. That is, 
\begin{align}
    \tilde{\vec{h}}^{l}_{u|p} = \tau(\vec{h}^{l-1}_u,
    \vec{s}^l_{p};\theta^l_\tau),
\end{align}
where the transformation function $\tau(\cdot;\theta^l_\tau)$ is learnable with parameters $\theta^l_\tau$. We implement $\tau$ using a layer of feature-wise linear modulation (FiLM) \cite{film,liu2021nodewise},
which enables the personalization of a message (\eg, $\vec{h}^{l-1}_u$) conditioned on arbitrary input (\eg, $\vec{s}^l_{p}$).
The FiLM layer learns to perform scaling and shifting operations to modulate the original message:
\begin{align}
    \tilde{\vec{h}}^{l}_{u|p} = (\gamma^{l}_{p}+\vec{1})\odot\vec{h}^{l-1}_u + \beta^{l}_{p},\label{eq.personalization}
\end{align}
where $\gamma^{l}_{p} \in \mathbb{R}^{d_h^{l-1}}$ is a scaling vector and $\beta^{l}_{p} \in \mathbb{R}^{d_h^{l-1}}$ is a shifting vector, both of which are learnable and specific to the path $p$.
Note that $\vec{1}$ is a vector of ones to center the scaling around one, and $\odot$ denotes the element-wise multiplication.
To make $\gamma^{l}_{p}$ and $\beta^{l}_{p}$ learnable, we materialize them using a fully connected layer, 
which takes in the semantic path embedding $\vec{s}^l_{p}$ as input to become conditioned on the path $p$, as follows.
\begin{align}
    \gamma^{l}_{p}=\textsc{LeakyReLU}(\vec{W}^{l}_{\gamma}\vec{s}^{l}_{p}+\vec{b}^{l}_{\gamma}),\label{eq.film-gamma} \\
    \beta^{l}_{p}=\textsc{LeakyReLU}(\vec{W}^{l}_{\beta}\vec{s}^{l}_{p}+\vec{b}^{l}_{\beta}),\label{eq.film-beta}
\end{align}
where $\vec{W}^{l}_{*}\in\mathbb{R}^{d^{l-1}_h \times d_s^{l}}$ and $\vec{b}^{l}_{*}\in\mathbb{R}^{d_h^{l-1}}$ are learnable parameters, and $\textsc{LeakyReLU}(\cdot)$ is the activation function. 
Note that the parameters of the transformation function $\tau$ in layer $l$ boil down to parameters of the fully connected layers, \ie, $\theta^l_\tau=\{\vec{W}^{l}_{\gamma},\vec{W}^{l}_{\beta},\vec{b}^{l}_{\gamma},\vec{b}^{l}_{\beta}\}$.

\stitle{Context aggregation.}
Next, we aggregate the personalized messages from latent heterogeneous contexts into $\vec{c}^{l}_v\in\mathbb{R}^{d_h^{l-1}}$,  the aggregated context embedding for the target node $v$: 
\begin{align} 
    \vec{c}^{l}_v = \textsc{Mean}(\{e^{-\lambda L(p)}\tilde{\vec{h}}^{l}_{u|p} \mid p\in P_v\}),\label{eq.aggr}
\end{align}
where $L(p)$ gives the length of the path $p$ so that $e^{-\lambda L(p)}$ acts
as a weighting scheme to bias toward shorter paths, and $\lambda>0$ is a hyperparameter controlling the decay rate. 
We use mean-pooling as the aggregation function, although other choices such as sum- or max-pooling could also be used.

Note that the self-information of the target node is also aggregated, by defining a
self-loop on the target node as a special path. More specifically, given a target node $v$ and its self-loop $p$, we define $L(p)=0$ and $\tilde{\vec{h}}^{l}_{v|p}$ = $\vec{h}^{l}_v$, which means the original message of the target node will be included into $\vec{c}_v^l$ with a weight of 1.

Finally, based on the aggregated context embedding, we obtain the primary embedding of node $v$ in the $l$-th layer:
\begin{align}
    \vec{h}^{l}_v = \textsc{LeakyReLU}(\vec{W}^{l}_h \vec{c}^{l}_v +\vec{b}^{l}_h),\label{eq.representation-update}
\end{align}
where $\vec{W}^{l}_h\in\mathbb{R}^{d_h^{l}\times d_h^{l-1}}$ and $\vec{b}^{l}_h\in\mathbb{R}^{d_h^{l}}$ are learnable parameters. 

\subsection{Link Prediction}

In the following, we discuss the treatment of the link prediction task on an LHG, as illustrated in Fig.~\ref{fig.model}(c). In particular, we will present a link encoder and the loss function.

\stitle{Link encoder.}
For link prediction between two nodes, we design a \emph{link encoder} to capture the potential latent relationships between the two candidates.
%
%
%
%
%
Given two candidate nodes $a$ and $b$ and their respective semantic embeddings $\vec{s}_a, \vec{s}_b$ obtained from the last \model\ layer, the link encoder is materialized in the form of a recurrent unit to generate a pairwise semantic embedding:
\begin{align}
    \vec{s}_{a,b} = \tanh{(\vec{W}\vec{s}_b + \vec{U}\vec{s}_a + \vec{b})},
\end{align}
where $\vec{s}_{a,b}\in\mathbb{R}^{d_h}$ can be interpreted as an embedding of the latent relationships between the two nodes $a$ and $b$, $\vec{W}, \vec{U}\in\mathbb{R}^{d_s\times d_h}$, $\vec{b}\in\mathbb{R}^{d_h}$ are learnable parameters. Here $d_h$ and $d_s$ are the number of dimensions of the primary and semantic embeddings from the last layer, respectively. Note that
$\vec{s}_{a,b}$ has the same dimension as the node representations, which can be used as a translation to relate nodes $a$ and $b$ in the loss function in the next part.


\stitle{Loss function.}
We adopt a triplet loss for link prediction. 
For an edge $(a,b) \in E$, we construct a triplet $(a,b,c)$, where  $c$ is a negative node randomly sampled from the graph. 
Inspired by translation models in knowledge graph \cite{transe, transr}, $b$ can be obtained by a translation on $a$ and the translation approximates the latent relationships between $a$ and $b$, \ie, 
$\vec{h}_b \approx \vec{h}_a + \vec{s}_{a,b}$. Note that $\vec{h}_v$ denotes the primary node embedding from the final \model\ layer.  
In contrast, since $c$ is a random node unrelated to $a$, $\vec{h}_c$ cannot be approximated by the translation. 
Thus, given a set of training triplets $T$, we formulate the following triplet margin loss for the task:
\begin{align}
    \bL_{\text{task}} =  \frac{1}{|T|}\sum_{(a,b,c)\in T} &\max\Big( d(a,b) - d(a,c)  + \alpha, 0 \Big), \label{eq:loss_task}
\end{align}
where $d(x,y)=\|\vec{h}_x + \vec{s}_{x,y}- \vec{h}_y\|_2$ is the Euclidean norm of the translational errors, 
and $\alpha > 0$ is the margin hyperparameter.







Besides the task loss, we also add constraints to scaling and shifting in the FiLM layer. 
During training, the scaling and shifting may become arbitrarily large to overfit the data. To prevent this issue, we restrict the search space by the following loss term on the scaling and shifting vectors.
\begin{align} 
    \bL_\text{FiLM}=\sum_{l=1}^{\ell}\sum_{p\in P}(\|\gamma^{l}_p\|_2 + \|\beta^{l}_p\|_2),\label{eq:loss_film}
\end{align}
where $\ell$ is the total number of layers and $P$ is the set of all sampled paths.
The overall loss is then
\begin{align} \label{eq.overall-loss}
    \bL = \bL_{\text{task}} + \mu \bL_\text{FiLM}, 
\end{align}
where 
$\mu>0$ 
is a hyperparameter to balance the loss terms.

We further present the training algorithm for \model\ in Appendix~\ref{sec.app.alg}, and give a complexity analysis therein.

\section{Experiments}
In this section, we conduct extensive experiments to evaluate the effectiveness of \model on four benchmark datasets.

\subsection{Experimental Setup}

\stitle{Datasets.} 
We employ four graph datasets summarized in Table~\ref{table.datasets}. Note that, while all these graphs include node or edge types, we hide such type information to transform them into LHGs.
%
%
\begin{itemize}[leftmargin=*]
\item \emph{FB15k-237} \cite{rotate} is a refined subset of Freebase \cite{freebase}, 
a knowledge graph about general facts. 
It is curated with the most frequently used relations, in which each node is an entity, and each edge represents a relation. 
%
\item \emph{WN18RR} \cite{rotate} is a refined subset of WordNet \cite{wordnet}, a knowledge graph demonstrating lexical relations of vocabulary. 
%
\item \emph{DBLP} \cite{gtn} is an academic bibliographic network, which includes three types of nodes, \ie, paper (P), author (A) and conference (C), as well as four types of edges, \ie, P-A, A-P, P-C, C-P. The node features are 334-dimensional vectors which represent the bag-of-word vectors for the keywords.
\item \emph{OGB-MAG} \cite{ogb} is a large-scale academic graph. It contains four types of nodes, \ie, paper (P), author (A), institution (I) and field  (F), as well as four types of edges, \ie, A-I, A-P, P-P, P-F. Features of each paper is a 128-dimensional vector generated by word2vec, while the node feature of the other types is generated by metapath2vec \cite{metapath2vec} with the same dimension following previous work \cite{pyg}. In our experiments, we randomly sample a subgraph with around 100K entities from the original graph using breadth first search.
\end{itemize}


\begin{table}[tbp]
\small
\caption{Summary of Datasets.\label{table.datasets}}
\begin{center}
\begin{tabular}{c|rrrr}
\toprule
\textbf{Attributes} & \textbf{FB15k-237} & \textbf{ WN18RR} & \textbf{DBLP} & \textbf{OGB-MAG} \\
\midrule
\# Nodes & 14,541 & 40,943 & 18,405 & 100,002  \\
\# Edges  & 310,116 & 93,003 & 67,946 & 1,862,256 \\
\# Features  & - & - & 334 & 128 \\
\# Node types  & - & - & 3 & 4 \\
\# Edge types  & 237 & 11 & 4 & 4 \\
Avg(degree)  & 29.09 & 3.50 & 3.55 & 17.88 \\ \midrule
\# Training  & 272,115 & 86,835 & 54,356 & 1,489,804 \\
\# Validation  & 17,535 & 3,034 & 6,794 & 186,225 \\
\# Testing  & 20,466 & 3,134 & 6796 & 186,227 \\
\bottomrule
\end{tabular}
\end{center}
\end{table}

\begin{table*}[!t]
    \centering
    \small
    \caption{Evaluation of link prediction on LHGs. Best is bolded and runner-up underlined; OOM means out-of-memory error.}
    \label{link_pred}
    \begin{tabular}{@{}l|cc|cc|cc|cc@{}}
    \toprule
  \multirow{2}*{Methods} & \multicolumn{2}{|c}{FB15k-237} & \multicolumn{2}{|c}{WN18RR} & \multicolumn{2}{|c}{DBLP} & \multicolumn{2}{|c}{OGB-MAG} \\
  
     & MAP & NDCG  & MAP & NDCG & MAP & NDCG & MAP & NDCG  \\
     \midrule
     
     GCN & 0.790 ± 0.001 & 0.842 ± 0.001 & 0.729 ± 0.002 & 0.794 ± 0.001 & 0.879 ± 0.001 & 0.910 ± 0.001 & 0.848 ± 0.001 & 0.886 ± 0.001 \\
    GAT & 0.786 ± 0.002 & 0.839 ± 0.001 & 0.761 ± 0.001 & 0.818 ± 0.001 &  \underline{0.913} ± 0.001 & \underline{0.936} ± 0.001 & 0.830 ± 0.004 & 0.872 ± 0.003 \\
    GraphSAGE & \underline{0.800} ± 0.001 & \underline{0.850} ± 0.001 & 0.728 ± 0.003 & 0.793 ± 0.002 & 0.891 ± 0.001 & 0.918 ± 0.001 & \underline{0.849} ± 0.001 & \underline{0.887} ± 0.001 \\
     \midrule
    TransE & 0.675 ± 0.001 & 0.752 ± 0.001 & 0.511 ± 0.002 & 0.624 ± 0.001 & 0.488 ± 0.001 & 0.605 ± 0.001 & 0.552 ± 0.001 & 0.656 ± 0.001 \\
    
    TransR & 0.734 ± 0.004 & 0.798 ± 0.003 & 0.510 ± 0.002 & 0.623 ± 0.001 & 0.565 ± 0.007 & 0.668 ± 0.005 & 0.546 ± 0.001 & 0.652 ± 0.001 \\
 
     \midrule
     HAN & 0.725 ± 0.002 & 0.793 ± 0.002 & 0.749 ± 0.003 & 0.810 ± 0.003 & 0.763 ± 0.005 & 0.801 ± 0.004 & OOM           & OOM           \\
  
    HGT & 0.782 ± 0.001 & 0.837 ± 0.001 & 0.724 ± 0.003 & 0.791 ± 0.002 & 0.897 ± 0.001 & 0.923 ± 0.001 & 0.835 ± 0.003 & 0.876 ± 0.002 \\
    
    HGN & 0.742 ± 0.002 & 0.806 ± 0.001 & \underline{0.802} ± 0.002 & \underline{0.849} ± 0.002 & 0.907 ± 0.003 & 0.930 ± 0.002 & 0.818 ± 0.001 & 0.863 ± 0.001 \\
   
     \midrule
     \model\ & \textbf{0.858} ± 0.001 & \textbf{0.893} ± 0.001 & \textbf{0.838} ± 0.003 & \textbf{0.877} ± 0.002 & \textbf{0.932} ± 0.003 & \textbf{0.949} ± 0.002 & \textbf{0.879} ± 0.001 & \textbf{0.909} ± 0.001 \\
     \bottomrule
     \end{tabular}
\end{table*}

\begin{table}
    \centering
    \small
     \addtolength{\tabcolsep}{-2pt}
    \caption{Evaluation of link prediction on LHGs with pseudo types for heterogeneous GNNs and translation models. 
    }
    \label{link_pred_pseudo}
    \begin{tabular}{@{}l|cc|cc|cc|cc@{}}
    \toprule
  \multirow{2}*{Methods} & \multicolumn{2}{|c}{FB15k-237} & \multicolumn{2}{|c}{WN18RR} & \multicolumn{2}{|c}{DBLP} & \multicolumn{2}{|c}{OGB-MAG} \\
  
     & MAP & NDCG  & MAP & NDCG & MAP & NDCG & MAP & NDCG  \\
     \midrule
     
    TransE-3 & 0.693 & 0.767 & 0.510 & 0.623 & 0.599 & 0.693 & 0.568 & 0.670  \\
    TransE-10 & 0.701  & 0.773  & 0.519  & 0.630  & 0.677 & 0.754 & 0.599  & 0.694  \\
    
    TransR-3 & 0.749 & 0.810 & 0.485  & 0.604 & 0.585  & 0.683 & 0.599  & 0.695  \\
    TransR-10 & 0.727 & 0.794 & 0.497 & 0.614 & 0.631 & 0.719 & OOM           & OOM           \\

    \midrule
    HAN-3 & 0.594 & 0.685  & 0.673 & 0.616  & 0.603 & 0.687  & OOM           & OOM           \\
    HAN-10 & 0.648  & 0.734  & 0.384  & 0.529  & 0.618  & 0.708 & OOM & OOM           \\
   
    HGT-3 & 0.799  & 0.850 & 0.733 & 0.797  & 0.888 & 0.916 & 0.837 & 0.878  \\
    HGT-10 & 0.750 & 0.812  & 0.607  & 0.701 & 0.857 & 0.893 &  0.837 & 0.878  \\
   
    HGN-3 & 0.746  & 0.809 & 0.814  & 0.859 & 0.903  & 0.927 & 0.815 & 0.861  \\
    
    HGN-10 & 0.735 & 0.800 & 0.822 & 0.864 & 0.898 & 0.923 & 0.813 & 0.859 \\

    \bottomrule
    \end{tabular}
\end{table}

\stitle{Baselines.} 
For a comprehensive comparison, we employ baselines from three major categories.
\begin{itemize}[leftmargin=*]
\item \emph{Classic GNNs}: GCN \cite{gcn}, GAT \cite{gat} and GraphSAGE \cite{sage}, which are classic GNN models for homogeneous graphs. 
\item \emph{Heterogeneous GNNs}: HAN \cite{han}, HGT \cite{hgt} and Simple-HGN (HGN for short) \cite{hgn}, which are state-of-the-art heterogeneous graph neural networks (HGNNs) taking in an HIN as input. 
\item \emph{Translation models}: TransE \cite{transe} and TransR \cite{transr}, which are well-known methods for knowledge graph embedding. 
\end{itemize}

Note that the heterogeneous GNNs and translation models require node/edge types as input, to apply them to an LHG, we adopt two strategies: either treating all nodes or edges as one type, or generating \emph{pseudo types},
which we will elaborate later. 
See  Appendix~\ref{sec.app.details} for more detailed descriptions of the baselines.

\stitle{Model settings.} See Appendix~\ref{sec.app.settings} for the hyperparameters and other settings of the baselines and our method.





\subsection{Evaluation of Link Prediction}
\label{sub_link}

In this part, we evaluate the performance of \model~on the main task of link prediction on LHGs.

\stitle{Settings.}
For knowledge graph datasets (FB15k-237 and WN18RR), we use the same training/validation/testing proportion in previous work \cite{rotate}, as given in Table~\ref{table.datasets}. For the other datasets, we adopt a 80\%/10\%/10\% random splitting of the links. 
Note that the training graphs are reconstructed from only the training links.


We adopt ranking-based evaluation metrics for link prediction, namely, NDCG and MAP \cite{tailgnn}. In the validation and testing set, given a ground-truth link $(a, b)$, we randomly sample another 9 nodes which are not linked to $a$ as negative nodes, and form a candidate list together with node $b$. For evaluation, we rank the 10 nodes based on their scores w.r.t.~node $a$. 
For our \model, the score for a candidate link $(x, y)$ is computed as $-\|\vec{h}_x + \vec{s}_{x,y}- \vec{h}_y\|_2$.
For classic and heterogeneous GNN models, we implement the same triplet margin loss for them, as given by Eq.~\eqref{eq:loss_task}. The only difference is that $d(x,y)$ is defined by $\|\vec{h}_x - \vec{h}_y\|_2$ in the absence of semantic embeddings. Similarly, their link scoring function is defined as $-\|\vec{h}_x - \vec{h}_y\|_2$ for a candidate link $(x,y)$. Translation models also use the same loss and scoring function as ours, except for replacing our link encoding $\vec{s}_{x,y}$ with their type-based relation embedding.

\stitle{Scenarios of comparison.} Since the type information is inaccessible on LHGs,
for heterogeneous GNNs and the translation methods, we consider the following scenarios.

The first scenario is to treat all nodes/edges as only one type in the absence of type information.

In the second scenario, we generate \emph{pseudo types}. For nodes, we resort to the $K$-means algorithm to cluster nodes into $K$ clusters based on their features, and treat the cluster ID of each node as its type. Since the number of clusters or types $K$ is unknown, we experiment with different values. For each heterogeneous GNN or translation model, we use ``X-$K$'' to denote a variant of model X with $K$ pseudo node types, where X is the model name. For instance, HAN-3 means HAN with three pseudo node types.  
Note that there is no node feature in FB15k-237 and WN18RR. To perform clustering, we use node embeddings by running X-1 first.
On the other hand, edge types are derived using the Cartesian product of the node types, resulting in $K\times K$ pseudo edge types. 
Finally, for HAN which requires metapath, we further construct pseudo metapaths based on the pseudo node types. For each pseudo node type, we employ all metapaths with length two starting and ending at that type. We also note that some previous works \cite{hovda2019missing, neelakantan2015missing} can predict missing type information. However, they cannot be used to generate the pseudo types, as they still need partial type information from some nodes and edges as supervision.

Besides, in the third scenario, we also evaluate the heterogeneous GNNs on a complete HIN with all node/edge types given. This assesses if explicit type information is useful, and how the baselines with full type information compare to our model.



\stitle{Performance on LHGs.}
In the first scenario, all methods do not have access to the node/edge types.
We report the results in Table~\ref{link_pred}, and make the following observations.

First, our proposed \model\ consistently outperforms all the baselines across different metrics and datasets. The results imply that \model\ can adeptly capture latent semantics between nodes to assist link prediction, even without any type information.

Second, the performance of classic GNN baselines is consistently competitive or even slightly better than heterogeneous GNNs. 
This finding is not surprising---while heterogeneous GNNs can be effective on HINs, their performance heavily depends on high-quality type information which is absent from LHGs.

Third, translation models are usually worse than GNNs, possibly because they do not take advantage of  node features, and lack a message-passing mechanism to fully exploit graph structures.

\stitle{Performance with pseudo types.}
In the second scenario, we generate pseudo types for heterogeneous GNNs and translation models, and report their results in Table~\ref{link_pred_pseudo}. 
We observe different outcomes on different kinds of baselines.

On one hand, translation models generally benefit from the use of pseudo types. Compared to Table~\ref{link_pred} without using pseudo types, TransE-$K$ can achieve an improvement of 13.2\% and 8.5\% in MAP and NDCG, respectively, while TransR-$K$ can improve the two metrics by 5.2\% and 3.6\%, respectively (numbers are averaged over the four datasets). This demonstrates that even very crude type estimation (\eg, $K$-means clustering) is useful in capturing latent semantics between nodes. Nevertheless, our model \model\ still outperforms translation models using pseudo types.

On the other hand, heterogeneous GNNs can only achieve marginal improvements with pseudo types, if not worse performance. A potential reason is that pseudo types are noisy, and the message-passing mechanism of GNNs can propagate local errors caused by the noises and further amplify them across the graph. In contrast, the lack of message passing in translation models make them less susceptible to noises, and the benefit of pseudo types outweighs the effect of noises.

Overall, while pseudo types can be useful to some extent, they cannot fully reveal the latent semantics between nodes due to potential noises. Moreover, we need to set a predetermined number of pseudo types, which is not required by our model \model.

\stitle{Performance on complete HINs.} 
The third scenario is designed to further evaluate the importance of type information, and how \model\ fares against baselines equipped with full type information on HINs. Specifically, we compare the performance of the heterogeneous GNNs on the two datasets DBLP and OGB-MAG, where type information is fully provided. To enhance the link prediction of the heterogeneous GNN models, we adopt a relation-aware decoder \cite{hgn} to compute the score for a candidate link $(x, y)$ as $\vec{h}_x^T \vec{W}_r \vec{h}_y$, where $\vec{W}_r \in \mathbb{R}^{d_h \times d_h}$ is a learnable matrix for each edge type $r \in R$. We report the results in Table~\ref{link_pred_true}.

We observe that heterogeneous GNNs with full type information consistently perform better than themselves without any type information (\emph{cf.} Table~\ref{link_pred}). This is not surprising given the rich semantics expressed by explicit types.
Moreover, \model\  achieves comparable results to the heterogeneous GNNs or sometimes better results (\emph{cf.} Table~\ref{link_pred}), despite \model\ not requiring any explicit type. A potential reason is the node- and path-level semantic embeddings in \model\ can capture latent semantics in a finer granularity, whereas the explicit types on a HIN may be coarse-grained. For example, on a typical academic graph, there are node types of Author or Paper, but no finer-grained types like Student/Faculty Author or Research/Application Paper is available.

\subsection{Evaluation of Node Type Classification} 

To evaluate the expressiveness of semantic embeddings in capturing type information, we further use them to conduct node type classification on DBLP and OGB-MAG thanks to their accessible ground truth types. 
We perform stratified train/test splitting, \ie, for each node type, we use 60\% nodes for training and 40\% for testing.
For each node, we concatenate its primary node embedding and semantic embedding, and feed it into a logistic regression classifier to predict its node type. We choose five competitive baselines, and use their output node embeddings to also train a logistic regression classifier on the same split.

We employ macro-F score and accuracy as the evaluation metrics, and report the results in Table~\ref{node_type}. We observe that \model\ can significantly outperform the other baselines, with only one exception in accuracy on OGB-MAG. Since the node types are imbalanced (\eg, authors account for 77.8\% of all nodes on DBLP, and 64.4\% on OGB-MAG), accuracy may be skewed by the majority class and is often not a useful indicator of predictive power. The results demonstrate the usefulness of semantic embedding in our model to effectively express type information.

  

\begin{table}[!t] 
    \centering
    \small
    \caption{Evaluation of link prediction on HINs with full access to node/edge types for heterogeneous GNNs. Percentages in parenthesis indicate the improvement to their performance on LHGs (\emph{cf.} Table~\ref{link_pred}).}
    \label{link_pred_true}
    \begin{tabular}{@{}l|cc|cc@{}}
    \toprule
    \multirow{2}*{Methods} & \multicolumn{2}{|c}{DBLP} & \multicolumn{2}{|c}{OGB-MAG} \\
  
     & MAP & NDCG & MAP & NDCG  \\
     \midrule
    HAN  & 0.789 (+3.4\%) & 0.821 (+2.5\%)  & OOM  & OOM       \\
    HGT  & 0.902 (+0.6\%) & 0.927 (+0.4\%) & 0.872 (+4.4\%) & 0.904 (+3.2\%)      \\
    HGN  & 0.909 (+0.2\%) & 0.932 (+0.2\%) & 0.855 (+4.5\%) & 0.892 (+3.4\%) \\
     \bottomrule
     \end{tabular}
\end{table}

\begin{table}
    \centering
    \small
    \addtolength{\tabcolsep}{-2pt}
    \caption{Evaluation of node type classification on LHGs.}
    \label{node_type}
    \begin{tabular}{@{}l|cc|cc@{}}
    \toprule
    \multirow{2}*{Methods} & \multicolumn{2}{|c}{DBLP} & \multicolumn{2}{|c}{OGB-MAG} \\
  
     & MacroF & Accuracy & MacroF & Accuracy  \\
     \midrule
     
    GCN  & 0.376 ± 0.009 & 0.785 ± 0.002 & 0.599 ± 0.011 & 0.890 ± 0.003 \\
    GAT  & 0.310 ± 0.003 & 0.782 ± 0.001 & 0.624 ± 0.035 & 0.894 ± 0.007 \\
    GraphSAGE & \underline{0.477} ± 0.021 & \underline{0.842} ± 0.012 & 0.550 ± 0.014 & 0.902 ± 0.004 \\
    \midrule
    HGT    & 0.464 ± 0.009 & 0.837 ± 0.005 & 0.823 ± 0.018 & \textbf{0.973} ± 0.003 \\
    HGN   & 0.292 ± 0.001 & 0.778 ± 0.001 & 0.531 ± 0.003 & 0.847 ± 0.003 \\
    \midrule
    LHGNN & \textbf{0.662} ± 0.001 & \textbf{0.995} ± 0.001 & \textbf{0.884} ± 0.002 & \underline{0.953} ± 0.001 \\
     \bottomrule
     \end{tabular}
\end{table}

\subsection{Model Analyses}


\begin{figure}[tbp]
\begin{minipage}{0.46\linewidth}
\addtolength{\tabcolsep}{-2pt}
\small
\captionof{table}{Training time.\label{table.time}}
\begin{tabular}{@{}rrrr@{}}
\toprule
\textbf{Nodes} & \textbf{Edges} & \textbf{Time} & \textbf{Epochs} \\
\midrule
20k & 370k & 1084s & 24 \\
40k & 810k & 1517s & 11 \\
60k & 1.2M & 2166s & 8 \\
80k & 1.6M & 2428s & 6 \\
100k & 1.8M & 2251s & 5 \\
\bottomrule
\end{tabular}
\end{minipage}%
\begin{minipage}{0.54\linewidth}
\centering
\includegraphics[width=0.99\linewidth]{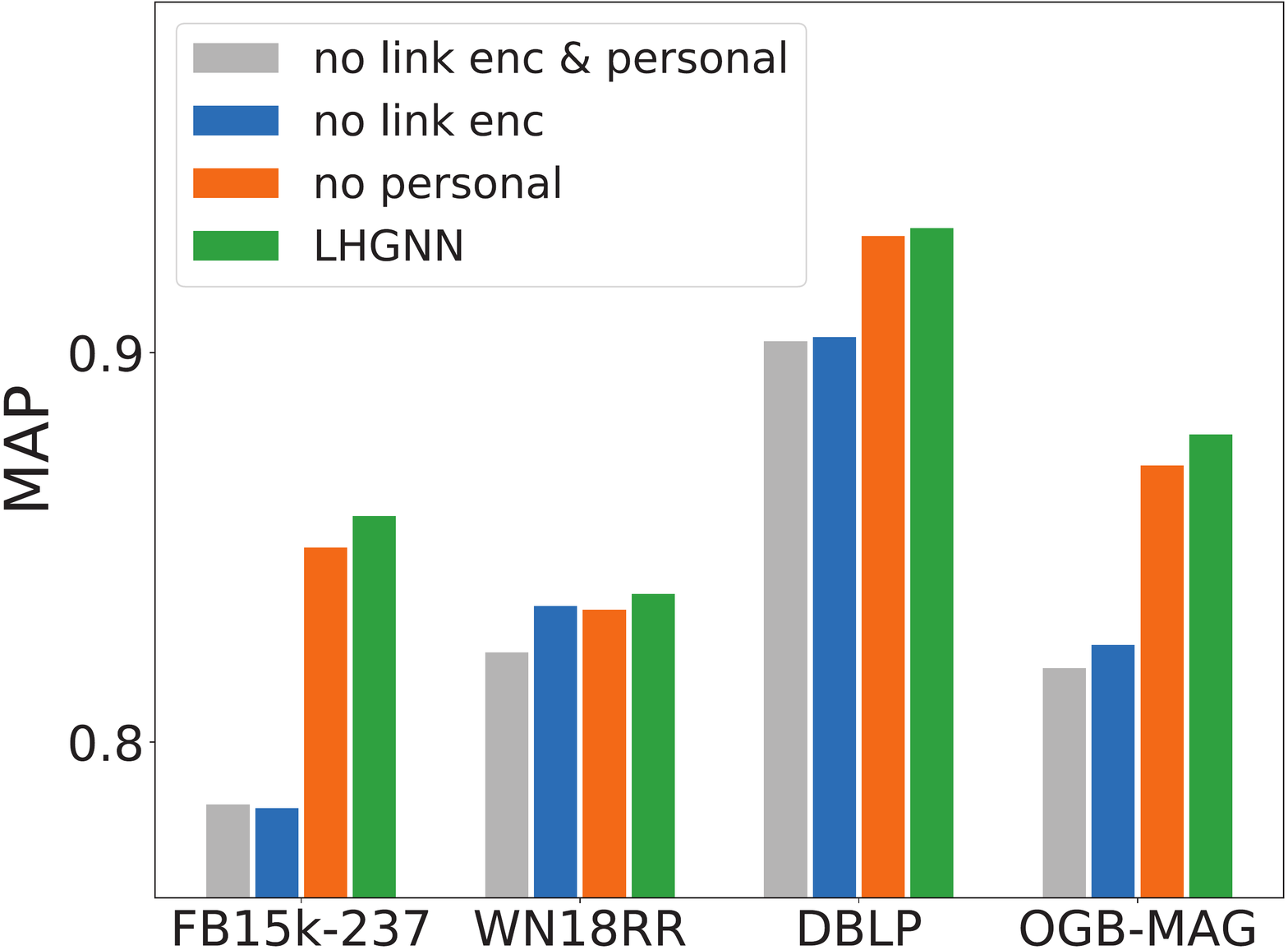}
\vspace{-6mm}
\captionof{figure}{Ablation study.}
\label{fig.ablation}
\vspace{-2mm}
\end{minipage}
\end{figure}

\stitle{Ablation study.}
To evaluate the contribution of each module in \model, we conduct an ablation study by comparing with several degenerate variants: 
(1) \emph{no link encoder}: we remove the link encoder, by setting all pairwise semantic embeddings $\vec{s}_{x,y}$ to zero in both training and testing;
(2) \emph{no personalization}: we remove the personalization conditioned on semantic path embeddings, by using a simple mean pooling for context aggregation; (3) \emph{no link encoder \& personalization}: we remove both modules as described earlier, which is equivalent to removing the semantic embeddings altogether.

We present the results in Fig.~\ref{fig.ablation} and make the following observations. 
First, the performance of \model\ drops significantly when removing the link encoder, showing their importance on LHGs. In other words, the learned latent semantics between nodes are effective for link prediction.
Second, without the personalization for context aggregation, the performance also declines. This shows that the context  nodes have heterogeneous relationships to the target node, and the semantic path embeddings can work as intended to personalize the contexts.
Third, without both of them, the model usually achieves the worst performance.

\stitle{Scalability.}
We sample five subgraphs from the largest dataset OGB-MAG, with sizes ranging from 20k to 100k nodes. We present the total training time and number of epochs of \model\ on these subgraphs in Table~\ref{table.time}. As the graph grows by 5 times, total training time to converge only increases by 2 times, since generally fewer epochs are needed for convergence on larger graphs. 


\stitle{Additional studies.} We present results on additional model studies in Appendices~\ref{sec.app.layers},~\ref{sec.app.size} and ~\ref{app:param_sensitivity}, respectively.




\section{Conclusion}

In this paper, we investigated a challenging and unexplored setting of latent heterogeneous graphs (LHG) for the task of link prediction. Existing approaches on heterogeneous graphs depend on explicit type-based information, and thus they do not work well on LHGs. 
To deal with the absence of types, we proposed a novel model named \model\ for link prediction on an LHG, based on the novel idea of semantic embedding at both node and path levels, and a personalized aggregation of latent heterogeneous contexts for target nodes in a fine-grained manner.
Finally, extensive experiments on four benchmark datasets show the superior performance of \model.

\begin{acks}
This research is supported by the Agency for Science, Technology
and Research (A*STAR) under its AME Programmatic Funds (Grant
No. A20H6b0151).
\end{acks}

\balance

\bibliographystyle{ACM-Reference-Format}
\bibliography{references}

\clearpage
\appendix
\section*{Appendices}
\renewcommand\thesubsection{\Alph{subsection}}
\renewcommand\thesubsubsection{\thesubsection.\arabic{subsection}}
\setcounter{secnumdepth}{5}
\renewcommand{\thesection}{\Alph{section}}
\renewcommand{\thefigure}{\Roman{figure}}
\renewcommand{\thetable}{\Roman{table}}

\subsection{Algorithm and Complexity} \label{sec.app.alg}
We outline the model training for \model\ in Algorithm~\ref{alg.train}.

\begin{algorithm}[h]
\small
\caption{\textsc{Model Training for} \model}
\label{alg.train}
\begin{algorithmic}[1]
    \Require latent heterogeneous graph $G=(V, E)$, training triplets $T$, a set of random walk paths $P_v$ for each node $v$
    \Ensure Model parameters $\Theta$. 
   \State initialize parameters $\Theta$;
    \While{not converged}
        \State sample a batch of triplets $T_{\text{bat}} \subset T$;
        \For{each target node $v$ in the batch $T_{\text{bat}}$}
            \For{each layer $ l \in\{1, 2, \ldots,\ell\}$}
            \State $\vec{s}^l_v \leftarrow f_s(\vec{h}^{l-1}_v;\theta^{l}_s)$; 
            \For{each path $p \in P_v$ that ends at context node $u$}
            \State $\vec{s}^{l}_{p} \leftarrow      f_p(\{\vec{s}_{v_i}^{l} | v_i \text{ in the path } p\})$;
            
            \State $\tilde{\vec{h}}^{l}_{u|p} \leftarrow \tau(\vec{h}^{l-1}_u,
             \vec{s}^l_{p};\theta^l_\tau) $;
            \EndFor
            
            \State $\vec{h}^{l}_v \leftarrow \text{Aggregate}(\{\tilde{\vec{h}}^{l}_{u|p} | p \in P_v\}$;
            \EndFor            
        \EndFor
        \State Calculate the loss $\bL$ by Eqs.~\eqref{eq:loss_task},\eqref{eq:loss_film} and \eqref{eq.overall-loss};
         
        \State update $\Theta$ by minimizing $\bL$;
    \EndWhile
    \State \Return $\Theta$.
\end{algorithmic}
\end{algorithm}

In line 1, we initialize the model parameters. In line 3, we sample a batch of triplets from training data. In lines 4--10, we calculate layer-wise node representations on the training set. Specifically, for each node $v$ in layer $l$, we calculate the semantic embeddings at the node level in line 6 and path level in line 8. Next,
we personalize the contexts in line 9 and aggregate them in line 10.
In lines 11-12, We compute the loss and update the parameters.


We compare the complexity of one layer and one target node in \model\ against a standard message-passing  GNN. In a standard  GNN, the aggregation for one node in the $l$-th layer has complexity $O(\bar{d}d_h^{l}d_h^{l-1})$, where $d_h^l$ is the output dimension of the $l$-th layer and $\bar{d}$ is the node degree. In our model, we first need to compute the node- and path-level semantic embeddings. 
At the node level, the cost is $O(d_s^{l}d_h^{l-1})$ based on Eq.~\eqref{eq:semantic-embedding-node};
at the path level, given a node with $k$ paths of maximum length $L_\text{max}$, the cost is $O(kL_\text{max} d^l_s)$ based on Eq.~\eqref{eq:semantic-embedding-path}. The computation of scaling and shifting vectors takes $O(kd_h^{l-1}d^{l}_s)$ based on Eqs.~\eqref{eq.film-gamma} and \eqref{eq.film-beta}, and the personalization of context embeddings takes $O(kd_h^{l-1})$ based on Eq.~\eqref{eq.personalization}. Thus, the aggregation and representation update step takes $O(kd_h^ld_h^{l-1})$ based on Eqs.~\eqref{eq.aggr} and \eqref{eq.representation-update}. 
Furthermore, the cost to sample a path of length $L_\text{max}$ is $O(L_\text{max})$. To sample $k$ paths for a target node in one LHGNN layer, the overhead is $O(kL_\text{max})$, which is negligible compared to the aggregation cost, as $L_\text{max}$ is small (less than 5 in our experiments). Therefore, the total complexity for one node in one layer is $O(kL_\text{max} d^l_s+kd_h^{l}d_h^{l-1} + kd_h^{l}d_s^{l-1})$. As  $L_\text{max}$ is a small constant and $d^{l-1}_s \ll d^{l-1}_h$, the complexity reduces to $O(kd_h^{l}d_h^{l-1})$. Furthermore, we can limit $k$, the number of sampled paths from each node, by some constant value, in which case our model has the same complexity class as a standard GNN.

\begin{table*}[!]
    \centering
    \small
    \caption{Impact of different number of layers. In each column, the best is bolded and the runner-up is underlined.}
    \label{num_layers}
    \begin{tabular}{@{}l|cc|cc|cc|cc@{}}
    \toprule
  \multirow{2}*{Methods} & \multicolumn{2}{|c}{FB15k-237} & \multicolumn{2}{|c}{WN18RR} & \multicolumn{2}{|c}{DBLP} & \multicolumn{2}{|c}{OGB-MAG} \\
  
     & MAP & NDCG  & MAP & NDCG & MAP & NDCG & MAP & NDCG  \\
     \midrule
     
     GCN-2 & \underline{0.790} ± 0.001 & \underline{0.842} ± 0.001 & 0.729 ± 0.002 & 0.794 ± 0.001 & 0.879 ± 0.001 & 0.910 ± 0.001 & 0.848 ± 0.001 & 0.886 ± 0.001 \\

    GCN-3 & 0.782 ± 0.001 & 0.837 ± 0.001 & 0.726 ± 0.005 & 0.792 ± 0.003 & 0.861 ± 0.001 & 0.896 ± 0.001 & 0.799 ± 0.003 & 0.849 ± 0.002 \\
    
   GCN-4 & 0.778 ± 0.001 & 0.833 ± 0.001 & 0.711 ± 0.005 & 0.781 ± 0.004 & 0.851 ± 0.003 & 0.888 ± 0.003 & 0.802 ± 0.002 & 0.851 ± 0.002 \\
     
     \midrule
    HGT-2 & 0.782 ± 0.001 & 0.837 ± 0.001 & 0.724 ± 0.003 & 0.791 ± 0.002 & 0.897 ± 0.001 & 0.923 ± 0.001 & 0.835 ± 0.003 & 0.876 ± 0.002 \\
    
    HGT-3 & 0.788 ± 0.001 & 0.841 ± 0.001 & 0.727 ± 0.012 & 0.792 ± 0.009 & 0.890 ± 0.001 & 0.917 ± 0.001 & 0.826 ± 0.002 & 0.870 ± 0.002 \\

    HGT-4 & 0.784 ± 0.005 & 0.838 ± 0.003 & 0.751 ± 0.007 & 0.811 ± 0.006 & 0.881 ± 0.011 & 0.911 ± 0.009 & 0.822 ± 0.003 & 0.867 ± 0.003 \\
   
     \midrule
     \model\ & \textbf{0.858} ± 0.001 & \textbf{0.893} ± 0.001 & \textbf{0.838} ± 0.003 & \textbf{0.877} ± 0.002 & \textbf{0.932} ± 0.003 & \textbf{0.949} ± 0.002 & \textbf{0.879} ± 0.001 & \textbf{0.909} ± 0.001 \\
     \bottomrule
     \end{tabular}
\end{table*}

\subsection{Details of Baselines}
\label{sec.app.details}

We provide detailed descriptions for the baselines. 

\begin{itemize}[leftmargin=*]
\item \emph{Classic GNNs}: GCN \cite{gcn} aggregates information for the target node by applying mean-pooling over its neighbors. GAT \cite{gat}: utilizes self-attention to assign different weights to neighbors of the target node during aggregation. Meanwhile, GraphSAGE \cite{sage}: concatenates the target node with the aggregated information from its neighbors to produce the node embedding. These models treat all nodes or edges as a uniform type and do not attempt to distinguish them.

\item \emph{Heterogeneous GNNs}:
HAN \cite{han} makes use of handcrafted metapaths to decompose HIN into multiple homogeneous graphs, one for each metapath, then employs hierarchical attention to learn both node-level and semantic-level importance for aggregation. 
HGT \cite{hgt} applies the transformer model, using node and edge type parameters to capture the heterogeneity. HGN \cite{hgn} extends GAT by employing node type information in the calculation of attention scores. These three models require type-based information in their architectures. Given an LHG, we either assume a single  node/edge type or employ pseudo types for these methods, as described in the main paper.

\item \emph{Translation models}:
TransE \cite{transe} models the relation between entities as a translation in the same embedding space. TransR \cite{transr} maps entity embeddings into a relation-wise space before the translation. These models require the relation type (i.e., edge type) to be known. Similarly, we assume a single edge type or employ pseudo types for them.

\end{itemize}

\subsection{Environment}
All experiments are conducted on a workstation with a 12-core CPU, 128GB RAM, and 2 RTX-A5000 GPUs. We implemented the proposed \model\ using  Pytorch 1.10 and Python 3.8 in Ubuntu-20.04.

\subsection{Model Settings}
\label{sec.app.settings}

For all the approaches, we use the same output dimension as 32 for fair comparison to conduct link prediction. For the two knowledge graph datasets (\ie, FB15k-237 and WN18RR), we randomly initialize a learnable parameter vector for each entity with embedding dimension 200. We tune the margin hyperparameter $\alpha$ for each model in order to achieve its optimal performance. All experiments are repeated 5 times, and we report the average results with standard deviations.

For all GNN baselines, we employ two layers with L2 Normalization on each layer. We use a margin $0.2$ and dropout ratio 0.5 for all of them. For GCN, we set the hidden dimension as 32.
For GAT, we use four attention heads with hidden dimension of each head as 16. For GraphSAGE, we use the mean aggregator and set its hidden dimension as 32.
For all HGNN baselines, we mainly follow the default setting in \cite{hgn}. In particular, we also employ 2-layer architectures for all of them. For HAN, for the node-level aggregation, we use GAT with eight attention heads with hidden dimension 8 for each head, and its dropout ratio is 0.6; and we set the dimension for semantic-level attention as 128 and set $\alpha=1$. For HGT, we use eight attention heads, with dropout ratio as 0.2 and margin as 1. For HGN, we use eight attention heads with dropout ratio as 0.5. We set the dimension of edge embeddings as 64, and the margin as 0.5.
For translation models including TransE and TransR, we set the learning rate as 0.005, embedding dimension as 200 and the margin as 0.2.

For our model, we also employ a two-layer architecture with L2-Normalization. We randomly sample 50 paths starting from each target node. We set the dimension of semantic embeddings as 10, decay ratio $\lambda$ as 0.1, the weight of scaling and shifting constraints $\mu$ as 0.0001, and margin as 0.2.

\subsection{Number of Layers}
\label{sec.app.layers}

We investigate the impact of number of layers in representative baseline models. Results from Table~\ref{num_layers} show that adding more layers might not give better results. Overall, they are still much worse than our proposed \model.

\subsection{Impact of Model Size}
\label{sec.app.size}

We investigate the impact of model size (\ie, the number of learnable parameters in a model) on empirical performance. Specifically, we select several representative GNNs from the baselines, and vary the size of each GNN by increasing its hidden dimension. For instance, GCN-32 indicates that its hidden layer has 32 neurons. 
Table~\ref{num_params} shows the link prediction performance of the GNNs with varying model sizes. Overall, larger models employing the same GNN architecture can only achieve slight improvements, and \model\ continues to outperform them despite having a relatively small model. The results imply that the effectiveness of \model\ comes from the architectural design rather than stacking with more parameters.

\begin{table}[t]
    \centering
    \small
     \addtolength{\tabcolsep}{-3pt}
    \caption{Impact of model size. $|\Theta|$ denotes the number of learnable parameters in a model.}
    \label{num_params}
    \resizebox{1\columnwidth}{!}{%
    \begin{tabular}{@{}l|rcc|rcc@{}}
    \toprule
    \multirow{2}*{Methods} & 
    \multicolumn{3}{|c|}{DBLP} &
    \multicolumn{3}{|c}{OGB-MAG} \\
    & $|\Theta|$ & MAP & NDCG & $|\Theta|$  &  MAP & NDCG\\
    
    \midrule
    GCN-32 & 12K & 0.879 ± 0.001 & 0.910 ± 0.001 & 5K & 0.848 ± 0.001 & 0.886 ± 0.001    \\
    GCN-64 & 25K & 0.890 ± 0.001 & 0.918 ± 0.001 & 12K & 0.849 ± 0.001 & 0.887 ± 0.001      \\
    GCN-96 & 41K & 0.890 ± 0.001 & 0.918 ± 0.001 & 21K & 0.847 ± 0.001 & 0.886 ± 0.001     \\
    
     \midrule
    GAT-16  & 26K & \underline{0.913} ± 0.001 & \underline{0.936} ± 0.001 & 13K & 0.830 ± 0.004 & 0.872 ± 0.003       \\
    GAT-32  & 60K & 0.910 ± 0.001  & 0.932 ± 0.001  & 33K & 0.828 ± 0.001  & 0.871 ± 0.001    \\
    GAT-48  & 102K & 0.910 ± 0.002 & 0.933 ± 0.002 & 62K & 0.815 ± 0.005 & 0.861 ± 0.004      \\

    \midrule
    HGT-32  & 21K & 0.897 ± 0.001 & 0.923 ± 0.001 & 14K & 0.835 ± 0.003 & 0.876 ± 0.002       \\
    HGT-64  & 61K & 0.907 ± 0.001 & 0.930 ± 0.001 & 48K & 0.836 ± 0.003  &  0.877 ± 0.002    \\
    
    \midrule
    HGN-32  & 239K & 0.907 ± 0.003 & 0.930 ± 0.002 & 231K & 0.818 ± 0.001 & 0.863 ± 0.001      \\
    HGN-64  & 717K & 0.905 ± 0.001 & 0.929 ± 0.001 & 703K & 0.826 ± 0.001 & 0.869 ± 0.001      \\
    
    \midrule
    LHGNN & 25K & \textbf{0.932} ± 0.003 & \textbf{0.949} ± 0.002 & 12K & \textbf{0.879} ± 0.001 & \textbf{0.909} ± 0.001 \\
     \bottomrule
     \end{tabular}}
\end{table}

\subsection{Parameter Sensitivity}\label{app:param_sensitivity}
\begin{figure}[t]
\centering
\includegraphics[width=0.99\linewidth]{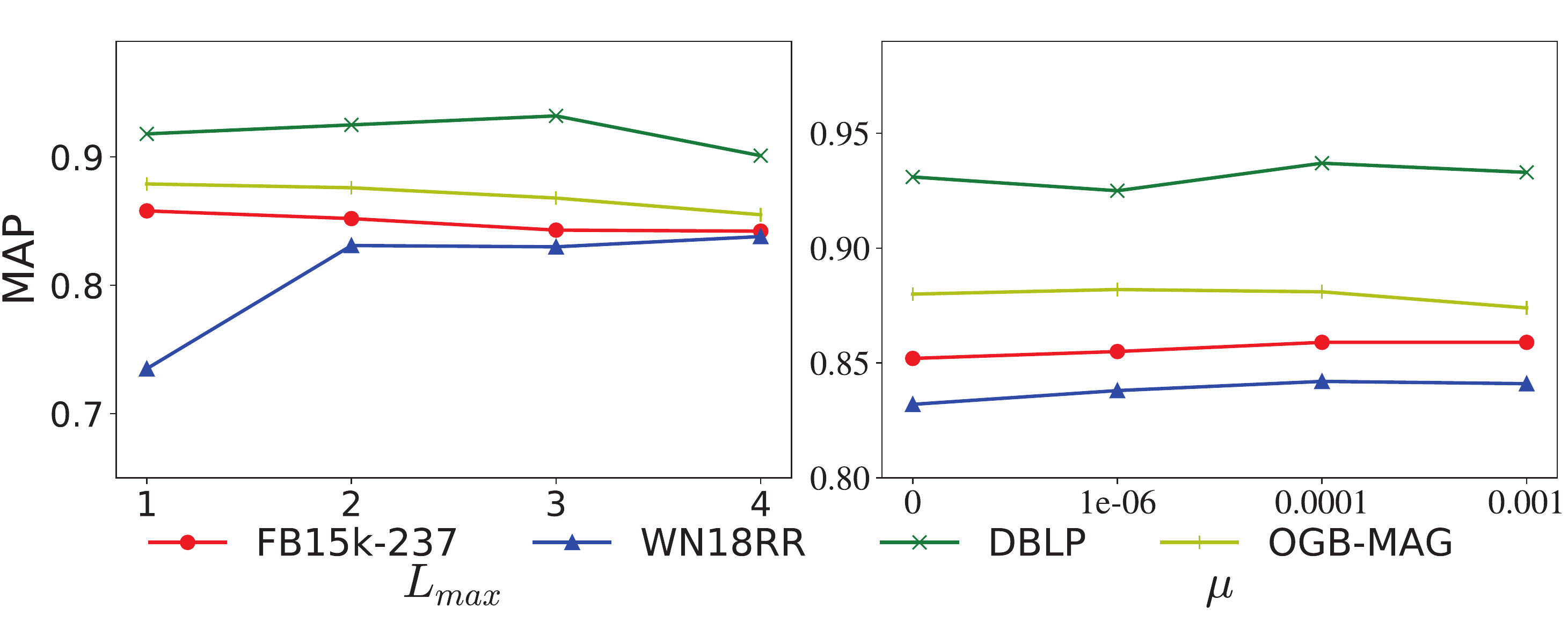}
\vspace{-2mm}
\caption{Impact of parameters.}
\label{fig.params}
\end{figure}

To study the impact of model parameters, we showcase two of the important parameters, including the maximum length for path sampling $L_{\text{max}}$, and the weight of scaling and shifting constraint $\mu$. We present the results in Fig.~\ref{fig.params}. 
For sparse datasets with low average degree such as WN18RR and DBLP, using a large maximum length for paths can generally improve the performance, as it can exploit more contextual structures around the target node. For the other two datasets, their performance is generally less affected as the maximum length of paths increases. 
For the hyper-parameter $\mu$, \model\  generally achieves the best performance in the interval [0.0001, 0.001] across the four datasets, demonstrating the necessity of this constraint.

\subsection{Data Ethics Statement}
To evaluate the efficacy of this
work, we conducted experiments which only use publicly available
datasets\footnote{\url{https://github.com/DeepGraphLearning/KnowledgeGraphEmbedding.git}}\footnote{\url{https://github.com/seongjunyun/Graph_Transformer_Networks.git}}\footnote{\url{https://github.com/snap-stanford/ogb.git}}, in accordance to their usage terms
and conditions if any.
We further declare that no personally identifiable information was
used, and no human or animal subject was involved in this research.

\end{document}